\documentclass[conference,a4paper]{IEEEtran}
\IEEEoverridecommandlockouts

\usepackage{cite}
\usepackage{amsmath,amssymb,amsfonts}
\usepackage{algorithmic}
\usepackage{graphicx}
\usepackage{textcomp}
\usepackage{xcolor}
\usepackage{multirow}

\def\BibTeX{{\rm B\kern-.05em{\sc i\kern-.025em b}\kern-.08em
    T\kern-.1667em\lower.7ex\hbox{E}\kern-.125emX}}

\begin{document}

\title{Affine-Aligned Atlas for Canonical Gaussian Construction in Video Representation}


\author{
\IEEEauthorblockN{Masaya Takabe}
 \IEEEauthorblockA{
    \textit{Graduate School of FSE,}\\
    \textit{Waseda University} \\
    Tokyo, Japan \\
    masaya.ta@asagi.waseda.jp}

\and
\IEEEauthorblockN{Hiroshi Watanabe}
  \IEEEauthorblockA{
    \textit{Graduate School of FSE,} \\
    \textit{Waseda University} \\
    Tokyo, Japan \\
    hiroshi.watanabe@waseda.jp}
\and
\IEEEauthorblockN{Sujun Hong}
 \IEEEauthorblockA{
    \textit{Sharp Corporation} \\
    Chiba, Japan \\
    hong.sujun@mail.sharp}
\and
\IEEEauthorblockN{Tomohiro Ikai}
 \IEEEauthorblockA{
    \textit{Sharp Corporation} \\
    Chiba, Japan \\
    ikai.tomohiro@mail.sharp}
}


\maketitle

\begin{abstract}
    Gaussian splatting has recently emerged as an efficient representation for images and videos due to its explicit structure and fast rendering capability. 
    Existing Gaussian-based video representations often decompose a video into canonical Gaussians and temporal deformation. 
    However, when a video contains large global motion such as camera movement, the canonical representation may become misaligned with individual frames, increasing the burden on the temporal deformation model.
    In this paper, we propose an affine-atlas canonical Gaussian representation, which constructs canonical Gaussians in a larger affine-aligned atlas space. 
    Frame-wise affine transforms absorb global motion before canonical Gaussian construction, reducing the gap between the canonical representation and target frames. 
    Since the proposed method only modifies the canonical construction stage, it can be integrated into existing canonical-Gaussian-based methods with negligible additional parameter cost. 
    Experiments show that our method improves reconstruction quality especially for sequences with large camera motion.
\end{abstract}


\section{Introduction}
    Video representation aims to express video data in a compact form while preserving reconstruction quality. 
    Neural video representations encode videos into network parameters~\cite{chen2021nerv,li2022nerv,chen2023hnerv, zhang2024boosting, yan2024ds, hayami2025sr}, but their decoding process requires repeated neural network inference. 
    Recently, Gaussian splatting has attracted attention as an explicit and efficient representation due to its fast rasterization capability~\cite{kerbl20233d}. This idea has also been extended to image and video representation~\cite{zhang2024gaussianimage,zhang2025image, tai2025eigengs, takabe2026contour, sun2024splatter, bond2025gaussianvideo, liu2025d2gv, lee2025gaussianvideo, pang2025gsvr, lin2026stgv}.    
    Compared with purely neural representations, Gaussian-based methods have explicit parameters and can be rendered efficiently, making them suitable for high-speed video decoding.

    \begin{figure}[t]
        \centering
        \includegraphics[width=0.98\linewidth]{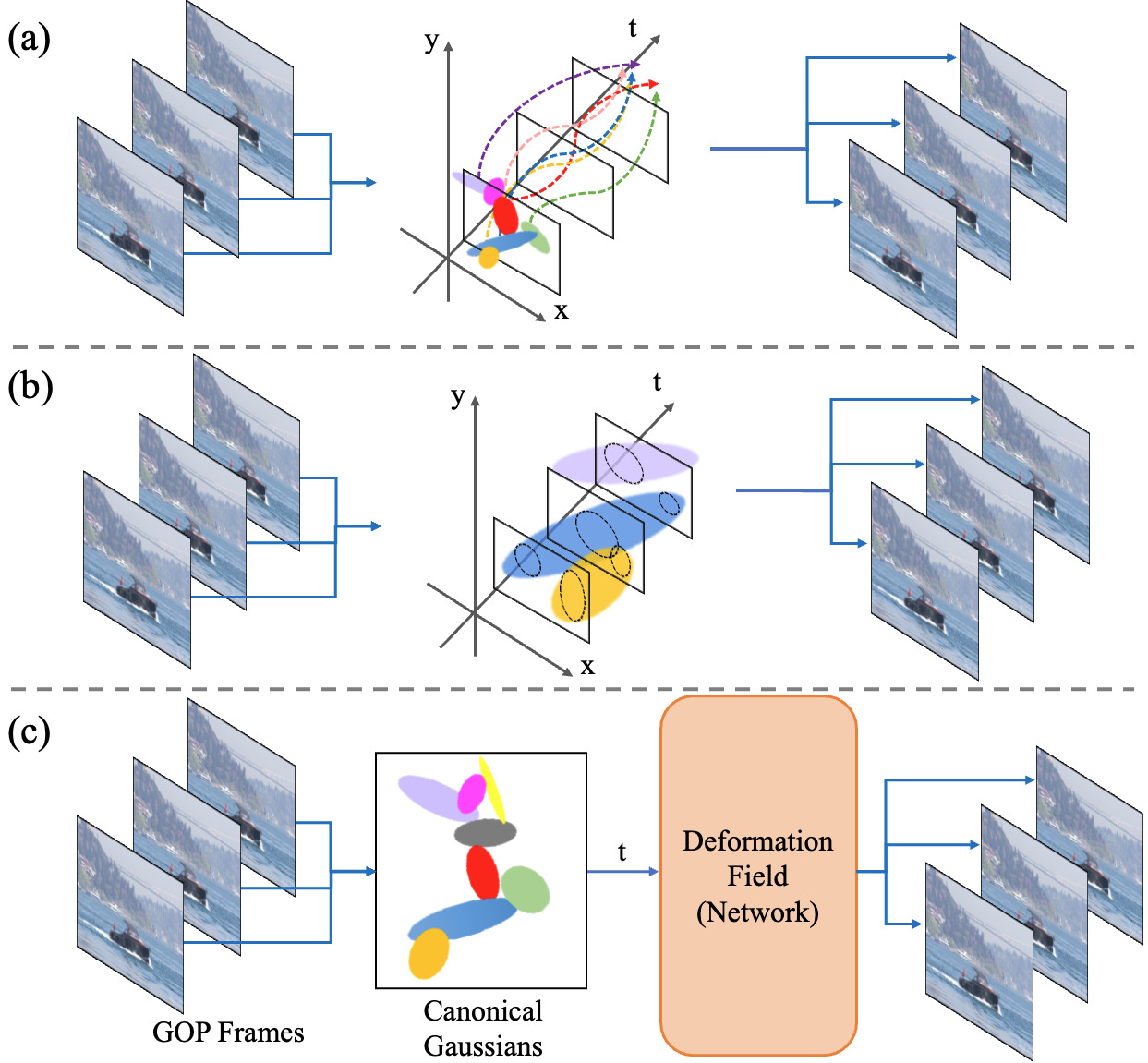}
        \caption{Categories of Gaussian-based video representation methods. 
        (a) Explicit temporal parameterization. 
        (b) Spatio-temporal Gaussian representation. 
        (c) Canonical Gaussians with temporal deformation. 
        Our work focuses on the canonical Gaussian construction stage in (c).}
        \label{fig:categories}
    \end{figure}

    Many recent Gaussian-based video representation methods adopt a canonical-Gaussian-based formulation, where canonical Gaussians are constructed for a short temporal segment, i.e., a group of pictures (GOP), and a temporal model predicts frame-dependent deformation. 
    This design is effective because it separates shared spatial structure from temporal variation, reducing redundant per-frame Gaussian parameters. 
    Therefore, the quality of the canonical Gaussians plays a central role in both reconstruction accuracy and representation efficiency.

\begin{figure*}[t]
\centerline{\includegraphics[width=0.95\textwidth]{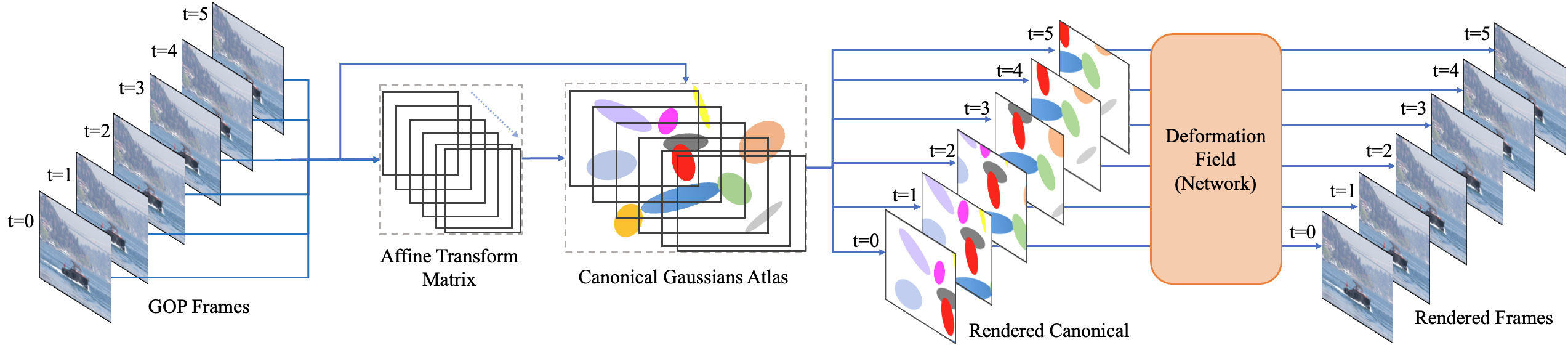}}
\caption{Overview of the proposed affine-atlas canonical Gaussian representation. GOP frames are aligned using frame-wise affine transforms, and canonical Gaussians are constructed in the atlas space. The atlas Gaussians are then rendered for each time step to obtain the rendered canonical representation.}
\label{fig:overview}
\end{figure*}

    However, when a video contains camera motion or large rigid motion, frames in the same GOP may not be well aligned. 
    The canonical Gaussians must then represent multiple shifted appearances in a single coordinate space, and the temporal deformation model must compensate for a large gap between the canonical representation and each target frame. 
    This can reduce representation efficiency and make temporal modeling more difficult.

    To address this issue, we propose an affine-atlas canonical Gaussian representation. 
    Instead of constructing canonical Gaussians directly in the original frame coordinate space, we introduce a larger atlas space and align GOP frames using frame-wise affine transforms. 
    By absorbing global motion during canonical Gaussian construction, the proposed method reduces the discrepancy between the rendered canonical representation and the target frames. 
    The proposed method only modifies the canonical construction stage and can therefore be integrated into existing canonical-Gaussian-based methods such as D2GV~\cite{liu2025d2gv}, GSVR~\cite{pang2025gsvr}, and STGV~\cite{lin2026stgv}. 
    In summary, our method improves canonical construction for motion-intensive sequences with negligible additional parameter cost.

\section{Related Work}
\subsection{Neural Video Representation}
    Neural video representation methods encode videos into neural network parameters and reconstruct frames through network inference~\cite{chen2021nerv,li2022nerv,chen2023hnerv, zhang2024boosting, yan2024ds, hayami2025sr}. 
    These methods demonstrate that videos can be represented compactly by exploiting spatial and temporal redundancy. 
    However, decoding requires repeated neural network inference for each frame, which can become a bottleneck when high decoding speed is required. 
    In contrast, Gaussian-based representations use explicit primitives and rasterization, which can provide faster rendering.

\subsection{Gaussian Splatting for Images and Videos}
    Gaussian splatting represents visual content using Gaussian primitives and enables efficient rasterization~\cite{kerbl20233d}. 
    This idea has been extended to compact 2D image representation~\cite{zhang2024gaussianimage,zhang2025image, tai2025eigengs, takabe2026contour}, video representation~\cite{sun2024splatter, bond2025gaussianvideo, liu2025d2gv, lee2025gaussianvideo, pang2025gsvr, lin2026stgv} and 4D representation~\cite{wu20244d, yang2024deformable, yang2024real, katsumata2024compact, li2024spacetime, luiten2024dynamic, chen2026pcgs}.

    As illustrated in Fig.~\ref{fig:categories}, Gaussian-based video representation methods can be categorized into three types according to how they model temporal variation.
    The first category explicitly parameterizes temporal changes of Gaussian attributes~\cite{sun2024splatter, bond2025gaussianvideo, katsumata2024compact}, which directly controls Gaussian motion but may require frame-dependent parameters or temporal coefficients.
    The second category represents videos using spatio-temporal Gaussian structures~\cite{smolak2025vegas,shah2026gaussian}, which directly model temporal continuity but can make representation and optimization more complex.
    The third category uses canonical Gaussians and temporal deformation~\cite{liu2025d2gv, pang2025gsvr, lin2026stgv, lee2025gaussianvideo}, where shared spatial structure is represented by canonical Gaussians and temporal variation is modeled by a deformation field or network. 
    This design is particularly attractive for video representation and compression because it separates shared spatial content from temporal variation.

\subsection{Canonical-Gaussian-Based Video Representation}
    Among Gaussian-based video representations, canonical-Gaussian-based methods are closely related to our work. 
    D2GV~\cite{liu2025d2gv} first fits 2D Gaussians to all frames in a GOP as a single canonical image and then learns temporal deformation using an MLP. 
    GSVR~\cite{pang2025gsvr} follows a similar strategy but uses a hybrid deformation field consisting of triplane grids and polynomial motion. 
    STGV~\cite{lin2026stgv} introduces spatio-temporal hash encoding for efficient temporal modeling.

    These methods demonstrate the effectiveness of the canonical Gaussian plus deformation framework. 
    However, when frames in a GOP are not well aligned due to global motion, the canonical space can become suboptimal, increasing the burden on temporal deformation. 
    Therefore, how to construct canonical Gaussians is a key factor for achieving efficient and accurate video representation.
    This motivates our affine-atlas formulation, which improves the canonical construction stage by absorbing global motion before temporal deformation.

\begin{table*}[t]
    \caption{
        Quantitative comparison on Big Buck Bunny and the UVG dataset. Each entry reports PSNR\(\uparrow\) / MS-SSIM\(\uparrow\). 
        UVG sequences are sorted from left to right by estimated affine motion magnitude, and the average is computed over UVG sequences. 
        Avg. \(\Delta\) denotes the average difference between each baseline and its +Ours variant over the three methods.
    }
    \label{tab:uvg_results}
    \centering
    \small
    \setlength{\tabcolsep}{0.63pt}
    \begin{tabular}{l|c|ccccccc|c}
        \hline
        \multirow{2}{*}{Method} & \multirow{2}{*}{BigBuckBunny} & \multicolumn{8}{c}{UVG} \\
        \cline{3-10}
        & & Jockey & ReadySetGo & YachtRide & Beauty & Bosphorus & HoneyBee & ShakeNDry & Average \\
        \hline\hline
        D2GV 
        & 36.63/0.9861 
        & 33.52/0.9218 
        & 27.53/0.8950 
        & 28.95/0.8952 
        & \textbf{33.82/0.9031} 
        & 32.89/0.9219 
        & 37.88/0.9836 
        & 34.53/0.9525 
        & 32.60/0.9226 \\

        +Ours 
        & \textbf{37.23/0.9885} 
        & \textbf{34.59/0.9391} 
        & \textbf{30.51/0.9468} 
        & \textbf{29.55/0.8982} 
        & 33.80/0.9028 
        & \textbf{33.87/0.9372} 
        & \textbf{37.90/0.9837} 
        & \textbf{34.70/0.9545} 
        & \textbf{33.47/0.9362} \\
        \hline

        GSVR 
        & 35.45/0.9796 
        & 27.01/0.8422 
        & 23.94/0.8208 
        & 26.90/\textbf{0.8558} 
        & 32.44/\textbf{0.8918} 
        & 33.05/0.9301 
        & 37.28/\textbf{0.9820} 
        & 32.51/0.9389 
        & 30.29/0.8911 \\

        +Ours 
        & \textbf{36.33/0.9851} 
        & \textbf{29.55}/\textbf{0.8836} 
        & \textbf{27.23}/\textbf{0.9095} 
        & \textbf{27.19}/0.8532 
        & \textbf{32.45}/0.8911 
        & \textbf{34.05}/\textbf{0.9513} 
        & \textbf{37.29}/\textbf{0.9820} 
        & \textbf{32.57}/\textbf{0.9392} 
        & \textbf{31.39}/\textbf{0.9139} \\
        \hline

        STGV 
        & \textbf{37.34/0.9888} 
        & 33.51/0.9221 
        & 27.54/0.8946 
        & 29.14/0.8938 
        & \textbf{33.85}/\textbf{0.9030} 
        & 33.35/0.9273 
        & 37.89/0.9837 
        & 34.80/\textbf{0.9561} 
        & 32.72/0.9235 \\

        +Ours 
        & 37.30/0.9885 
        & \textbf{34.64}/\textbf{0.9394} 
        & \textbf{30.81}/\textbf{0.9499} 
        & \textbf{29.56}/\textbf{0.8962} 
        & 33.81/0.9025 
        & \textbf{33.90}/\textbf{0.9368} 
        & \textbf{37.92}/\textbf{0.9838} 
        & \textbf{34.82}/0.9557 
        & \textbf{33.50}/\textbf{0.9425} \\
        \hline

        Avg. $\Delta$
        & +0.48/+0.0025
        & +1.58/+0.0253
        & +3.18/+0.0653
        & +0.44/+0.0009
        & -0.02/-0.0005
        & +0.84/+0.0153
        & +0.02/+0.0001
        & +0.08/+0.0006
        & +0.92/+0.0185 \\
        \hline
    \end{tabular}
\end{table*}

\section{Proposed Method}
\subsection{Overview}
    We propose an affine-atlas canonical Gaussian representation for efficient video representation. As shown in Fig.~\ref{fig:overview}, the proposed method constructs canonical Gaussians in a larger atlas space instead of directly optimizing them in the original frame coordinate space. Each frame in a GOP is associated with a frame-wise 6-parameter affine transform, and the frames are aligned in the atlas space before canonical Gaussian construction.
    This design allows global motion, such as translation, rotation, and scaling, to be absorbed before canonical Gaussian construction.
    After this construction stage, each baseline temporal deformation model is applied without modification.

\subsection{Affine-Atlas Canonical Space}
    Let a GOP contain $T$ frames $\{I_t\}_{t=0}^{T-1}$. 
    Conventional canonical-Gaussian-based methods construct a canonical Gaussian set directly from these frames or a single key frame in the original frame coordinate space. 
    In contrast, we define a larger atlas space that covers the affine-aligned regions of all frames in the GOP.
    Each frame $I_t$ is associated with an affine transform matrix $\mathbf{A}_t$, which maps coordinates in the atlas space to coordinates in the frame space.
    These affine transforms are precomputed, fixed during optimization, and used to project atlas Gaussian positions into each target frame coordinate system.
    The canonical Gaussian set in the atlas space is represented as
    \begin{equation}
        \mathcal{G}_{\mathrm{atlas}} =
        \{(\boldsymbol{\mu}_i, \mathbf{\Sigma}_i, \mathbf{c}_i)\}_{i=1}^{N},
        \label{eq:atlas_gaussians}
    \end{equation}
    where $\boldsymbol{\mu}_i$ denotes the center position of Gaussians in the atlas space, $\mathbf{\Sigma}_i$ denotes its covariance matrix, $\mathbf{c}_i$ denotes color, and $N$ is the number of Gaussians.

    For each frame $t$, we define a frame-wise canonical Gaussian set projected from the atlas as
    \begin{equation}
        \mathcal{G}_{\mathrm{atlas}}^t =
        \{({\boldsymbol{\mu}}_{i}^t, \mathbf{\Sigma}_i, \mathbf{c}_i)\}_{i=1}^{N}.
    \end{equation}
    The projected position is computed by the precomputed affine transform
    \begin{equation}
         {\boldsymbol{\mu}}_{i}^t =\mathbf{L}_t \boldsymbol{\mu}_i + \mathbf{b}_t,
    \end{equation}
    where \(\mathbf{A}_t = [\mathbf{L}_t \mid \mathbf{b}_t]\) maps the atlas coordinates to the frame coordinates.
    In the implementation, the affine transform is applied only to Gaussian positions, while the covariance and color are shared from the atlas Gaussian set.
    By using this transform, the same atlas Gaussian can more consistently explain corresponding regions across multiple frames under global motion.

\subsection{Canonical Gaussian Construction}
    Given the precomputed affine transforms, the objective of canonical Gaussian construction is to optimize the atlas Gaussian parameters so that the rendered images match the target GOP frames after the affine transform. 
    Let $\mathcal{R}(\cdot)$ denote the Gaussian rasterizer. For frame $t$, the rendered frame is obtained as
    \begin{equation}
        \hat{I}_t = \mathcal{R}(\mathcal{G}_{\mathrm{atlas}}^t).
        \label{eq:rendering}
    \end{equation}
    The reconstruction loss is defined as
    \begin{equation}
        \mathcal{L}_{\mathrm{rec}} = \sum_{t=0}^{T-1} d(\hat{I}_t, I_t),
        \label{eq:reconstruction_loss}
    \end{equation}
    where $d(\cdot,\cdot)$ is an image reconstruction loss such as an L2 loss. 
    Through this optimization, the atlas Gaussians are encouraged to represent the shared content of the GOP in an aligned coordinate system.

    During the following temporal modeling stage, $\mathcal{G}_{\mathrm{atlas}}^t$ is further deformed by the baseline deformation model before final rendering.
    Compared with conventional canonical construction, the proposed formulation reduces the gap between the canonical representation and each target frame. 
    Since global motion is handled by affine transforms, the subsequent temporal deformation model can focus more on residual local motion and appearance changes.

\begin{figure*}[t]
    \centerline{\includegraphics[width=0.95\textwidth]{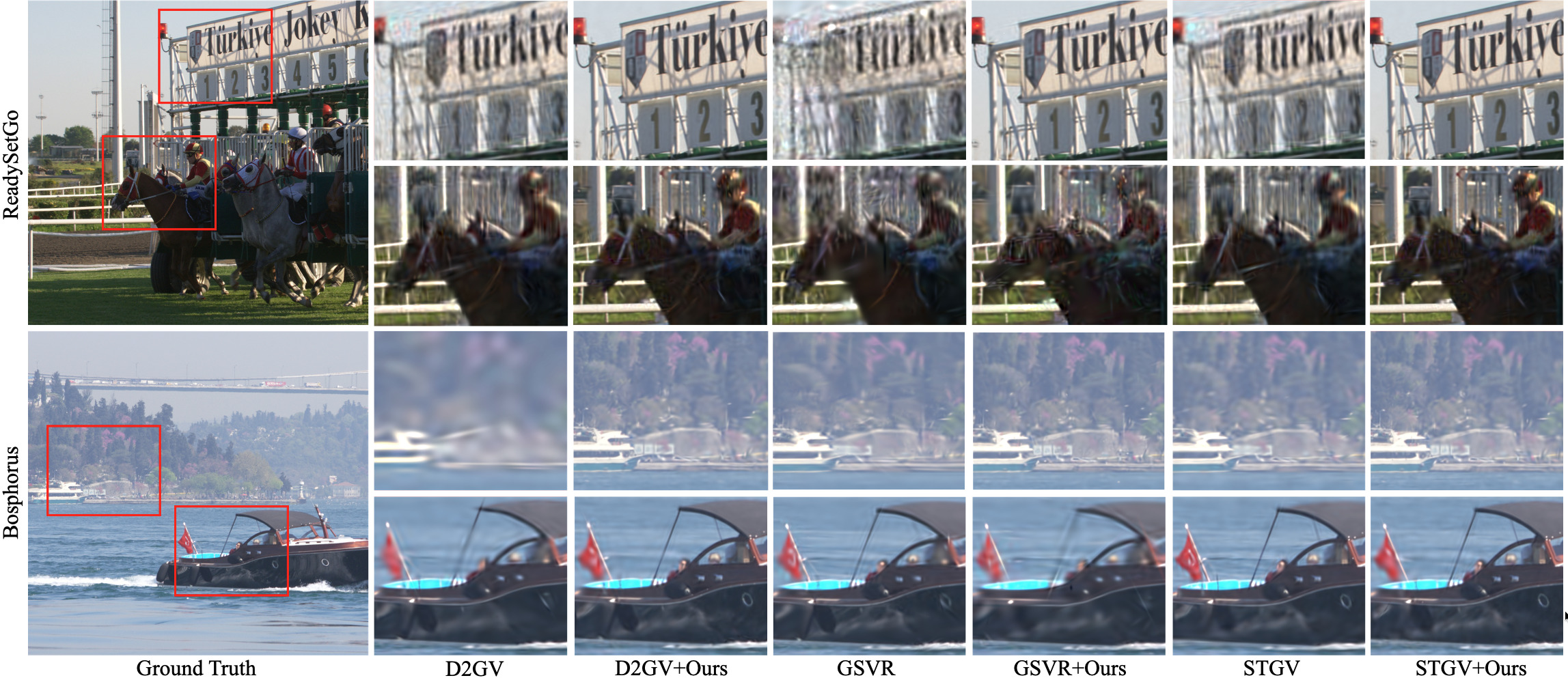}}
    \caption{Qualitative comparison on UVG sequences with large camera motion. The proposed affine-atlas construction improves the reconstruction of regions affected by global camera motion.}
    \label{fig:qualitative}
\end{figure*}

\begin{figure*}[t]
    \centerline{\includegraphics[width=0.95\textwidth]{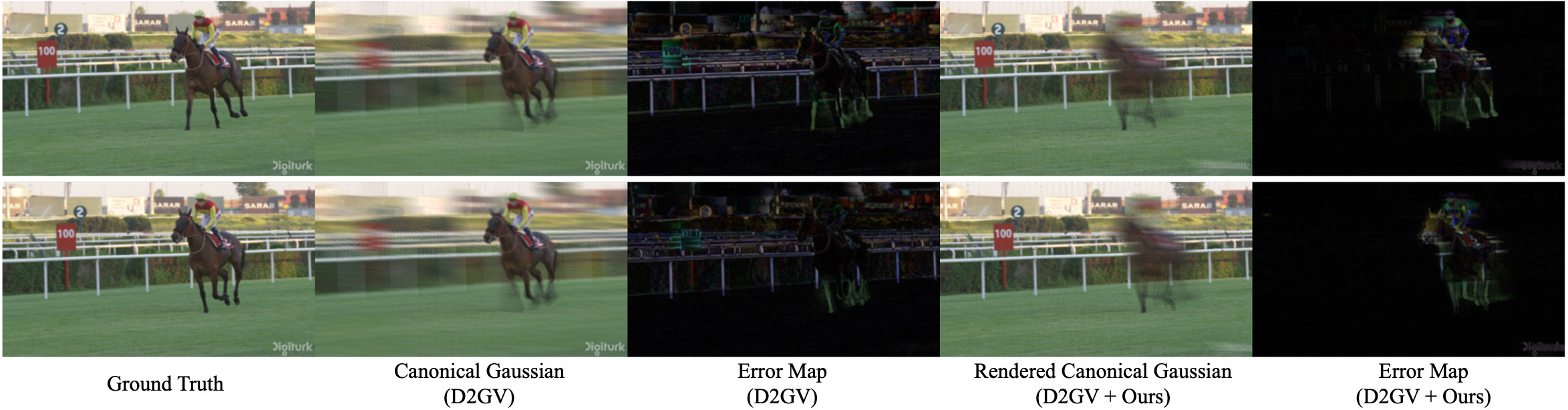}}
    \caption{Qualitative comparison of canonical Gaussian construction on a sequence with large camera motion. The proposed affine-atlas construction reduces the gap between the canonical representation and the target frames.}
    \label{fig:canonical_diff}
\end{figure*}

\section{Experiments}
\subsection{Experimental Setup}
    We conduct experiments on Big Buck Bunny~\cite{roosendaal2008big} and the UVG dataset~\cite{mercat2020uvg}. 
    For Big Buck Bunny, we use 132 frames with a resolution of $1280 \times 720$. For the UVG dataset, we use a spatial resolution of $1920 \times 1080$ and temporally subsample each sequence by a factor of five. 
    Each video sequence is divided into GOPs of 10 frames, and each GOP is reconstructed independently. 
    This setting follows the common video representation pipeline, where canonical Gaussians and temporal deformation are optimized for each GOP.
    We evaluate the reconstruction quality using PSNR and MS-SSIM. 
    We compare each baseline method with its affine-atlas variant: D2GV, GSVR, and STGV are evaluated with and without the proposed canonical Gaussian construction. 
    In all variants, only the canonical construction stage is replaced, while the temporal modeling component of each method is kept unchanged.

    For the affine-atlas variants, frame-wise affine transforms are precomputed using ORB feature matching. 
    We estimate a two-dimensional affine transform between adjacent frames and compose them to obtain frame-wise transforms within each GOP. 
    The estimated affine matrices are fixed during training and are not optimized as learnable parameters.
    We also use the average translation magnitude of the adjacent-frame affine transforms as the global motion magnitude for sorting UVG sequences in Table~\ref{tab:uvg_results}.

\subsection{Results and Analysis}
    Table~\ref{tab:uvg_results} shows the quantitative comparison on Big Buck Bunny and the UVG dataset.
    The proposed affine-atlas construction improves reconstruction quality, especially for sequences with large camera motion, such as Bosphorus, Jockey, and ReadySetGo. 
    This tendency is observed across the evaluated baseline methods, suggesting that the proposed method improves the canonical Gaussian construction stage rather than a specific temporal deformation model.

    In contrast, the improvement is limited for sequences with relatively small motion, such as Big Buck Bunny, Beauty and HoneyBee. 
    This is because the conventional canonical Gaussians and the affine-atlas canonical Gaussians become similar when there is little global motion. 
    In such cases, the original canonical space is already well aligned with the target frames, and the benefit of affine alignment becomes small.

    Fig.~\ref{fig:qualitative} shows qualitative reconstruction results on UVG sequences with large camera motion. 
    The affine-atlas variants better preserve local structures in regions affected by camera motion, such as background signs and moving objects. 
    Fig.~\ref{fig:canonical_diff} further compares the rendered canonical Gaussians and their error maps. 
    The baseline canonical representation becomes blurred because misaligned frames are fitted in a single coordinate space, whereas the proposed affine-atlas construction reduces this discrepancy by aligning frames before canonical Gaussian optimization.

\subsection{Parameter Cost}
    The proposed method introduces only frame-wise affine transform matrices and does not add a neural network or modify the temporal deformation model.
    Since a two-dimensional affine transform is represented by a $2 \times 3$ matrix, the additional cost is only $6T$ parameters for a GOP with $T$ frames. 
    For example, a 10-frame GOP requires only $60$ affine parameters, which is negligible compared with the Gaussian and temporal-model parameters, typically on the order of hundreds of thousands per GOP.
    This suggests that the performance gain comes from better canonical alignment, not from simply increasing the model size.

\section{Conclusion}
    In this paper, we propose an affine-aligned atlas for canonical Gaussian construction in video representation. 
    By constructing canonical Gaussians in a larger affine-aligned atlas space, the proposed method absorbs global motion before temporal deformation learning and reduces the gap between the canonical representation and target frames. 
    Experiments on Big Buck Bunny and UVG sequences show that our method improves reconstruction quality, especially for sequences with large camera motion, with negligible parameter cost.
    Future work includes more robust estimation of alignment transforms and extending the atlas construction to more flexible transformations beyond affine motion.

\bibliographystyle{IEEEtran}
\bibliography{reference}

\end{document}